\documentclass{article}
\usepackage{microtype}
\usepackage{graphicx}
\usepackage{subfigure}
\usepackage{booktabs}
\usepackage[nohyperref,accepted]{icml2025}

\usepackage{algorithm}
\usepackage{algorithmicx}
\usepackage[noend]{algpseudocode}
\usepackage{amsfonts}
\usepackage{amsmath}
\usepackage{amssymb}
\usepackage{amsthm}
\usepackage{bbm}
\usepackage{bm}
\usepackage{color}
\usepackage{dirtytalk}
\usepackage{dsfont}
\usepackage{enumerate}
\usepackage{listings}
\usepackage{mathtools}
\usepackage{natbib}
\usepackage{times}
\usepackage{xspace}

\usepackage[dvipsnames]{xcolor}
\usepackage[bookmarks=false]{hyperref}
\hypersetup{
  pdffitwindow=true,
  pdfstartview={FitH},
  pdfnewwindow=true,
  colorlinks,
  linktocpage=true,
  urlcolor=Green,
  linkcolor=Green,
  citecolor=Green
}
\usepackage[capitalize,noabbrev]{cleveref}

\usepackage[textsize=tiny]{todonotes}

\usepackage{thmtools}
\declaretheorem[name=Theorem,refname={Theorem,Theorems},Refname={Theorem,Theorems}]{theorem}
\declaretheorem[name=Lemma,refname={Lemma,Lemmas},Refname={Lemma,Lemmas},sibling=theorem]{lemma}

\declaretheorem[name=Assumption,refname={Assumption,Assumptions},Refname={Assumption,Assumptions}]{assumption}

\newcommand{\cD}{\mathcal{D}}
\newcommand{\cE}{\mathcal{E}}

\newcommand{\cL}{\mathcal{L}}

\newcommand{\cN}{\mathcal{N}}

\newcommand{\cS}{\mathcal{S}}

\newcommand{\cX}{\mathcal{X}}
\newcommand{\cY}{\mathcal{Y}}
\newcommand{\cZ}{\mathcal{Z}}

\newcommand{\realset}{\mathbb{R}}

\newcommand{\E}[1]{\mathbb{E}\left[#1\right]}

\newcommand{\Erv}[2]{\mathbb{E}_{#1}\left[#2\right]}

\newcommand{\abs}[1]{\left|#1\right|}

\newcommand{\I}[1]{\mathds{1} \! \left\{#1\right\}}

\newcommand{\norm}[1]{\|#1\|}
\newcommand{\normw}[2]{\|#1\|_{#2}}

\newcommand{\set}[1]{\left\{#1\right\}}

\newcommand{\T}{^\top}

\DeclareMathOperator*{\argmax}{arg\,max\,}
\DeclareMathOperator*{\argmin}{arg\,min\,}
\let\det\relax
\DeclareMathOperator{\det}{det}

\let\trace\relax
\DeclareMathOperator{\trace}{tr}
\mathchardef\mhyphen="2D

\newcommand{\adpo}{\ensuremath{\color{Green}\tt ADPO}\xspace}
\newcommand{\adpoplus}{\ensuremath{\color{Green}\tt ADPO^+}\xspace}
\newcommand{\apo}{\ensuremath{\color{Green}\tt APO}\xspace}
\newcommand{\pmc}{\ensuremath{\color{Green}\tt PMC}\xspace}
\newcommand{\uniform}{\ensuremath{\color{Green}\tt Uniform}\xspace}

\icmltitlerunning{Active Learning for Direct Preference Optimization}

\begin{document}

\twocolumn[
\icmltitle{Active Learning for Direct Preference Optimization}

\icmlsetsymbol{equal}{*}

\begin{icmlauthorlist}
\icmlauthor{Branislav Kveton}{ar}
\icmlauthor{Xintong Li}{ucsd}
\icmlauthor{Julian McAuley}{ucsd}
\icmlauthor{Ryan Rossi}{ar}
\icmlauthor{Jingbo Shang}{ucsd}
\icmlauthor{Junda Wu}{ucsd}
\icmlauthor{Tong Yu}{ar}
\end{icmlauthorlist}

\icmlaffiliation{ar}{Adobe Research}
\icmlaffiliation{ucsd}{University of California, San Diego}

\icmlcorrespondingauthor{Branislav Kveton}{kveton@adobe.com}

\vskip 0.3in
]

\printAffiliationsAndNotice{}

\begin{abstract}
Direct preference optimization (DPO) is a form of reinforcement learning from human feedback (RLHF) where the policy is learned directly from preferential feedback. Although many models of human preferences exist, the critical task of selecting the most informative feedback for training them is under-explored. We propose an active learning framework for DPO, which can be applied to collect human feedback online or to choose the most informative subset of already collected feedback offline. We propose efficient algorithms for both settings. The key idea is to linearize the DPO objective at the last layer of the neural network representation of the optimized policy and then compute the D-optimal design to collect preferential feedback. We prove that the errors in our DPO logit estimates diminish with more feedback. We show the effectiveness of our algorithms empirically in the setting that matches our theory and also on large language models.
\end{abstract}

\section{Introduction}
\label{sec:introduction}

\emph{Reinforcement learning from human feedback (RLHF)} has been effective in aligning and fine-tuning \emph{large language models (LLMs)} \citep{ouyang22training,rafailov23direct}. The main difference from classic \emph{reinforcement learning (RL)} \citep{sutton98reinforcement} is that the agent learns from human feedback, which is expressed as preferences for different potential choices \citep{christiano17deep}. The human feedback allows LLMs to be adapted beyond the distribution of data that was used for their pre-training and generate more human-like responses. The feedback can be incorporated by learning a reward model \citep{ouyang22training} from preferences over two \citep{bradley52rank} or multiple \citep{plackett75analysis,luce05individual} choices. \emph{Proximal policy optimization (PPO)} \citep{schulman17proximal} is then used to maximize the expected reward of the LLM policy under the reward model. Learning of reward models can be avoided by directly optimizing the policy with preferential feedback, known as \emph{direct preference optimization (DPO)} \citep{rafailov23direct}.

Learning of human preferences for LLM optimization has two main components: preference modeling \citep{rafailov23direct,ethayarajh24model} and how the preferences are elicited \citep{lightman24lets}. We focus on the latter and note that this problem is analogous to classic active learning \cite{bishop06pattern}. Prior works formulated this problem as identifying a subset of prompts with candidate responses, either online or offline, where preferential feedback would improve policy learning by RLHF, either through a reward model or DPO. These works differ in how the prompts are selected: \citet{mehta23sample,ji2024reinforcement,muldrew2024active} choose prompts based on differences of estimated rewards to their responses; \citet{mukherjee24optimal,scheid24optimal,thekumparampil24comparing} derive optimal policies for offline exploration using D-optimal designs \citep{pukelsheim06optimal}; and \citet{das24active,liu24dual} solve D-optimal designs online using a greedy algorithm. Most works prove that the errors in learned reward models diminish with more feedback. Interestingly, many works propose two kinds of algorithms \citep{mehta23sample,das24active,ji2024reinforcement}, which are either analyzable or practical. We present the first analysis of active learning in DPO and our algorithms are practical.

We study active learning in direct preference optimization. At a high level, we collect preferential feedback to improve DPO policies learned from it. We study two settings: online and offline. In the \emph{online setting}, the input is a dataset of $N$ prompts with two candidate responses per prompt. The human feedback is unknown in advance and we elicit it online. This setting is motivated by statistical efficiency; we elicit the most informative feedback within a fixed budget on human labor. In the \emph{offline setting}, the input is a dataset of $N$ prompts with two candidate responses per prompt, and logged preferential feedback for the responses. This setting is motivated by computational efficiency; even if the human feedback is known in advance, we may not have computational resources to learn from all of it. We solve both settings in a unified way. The key idea in our work is to linearize the DPO objective at the last layer of the neural network representation of the optimized policy and identify the most informative subset of $n$ prompts out of $N$ using a D-optimal design \citep{pukelsheim06optimal}. D-optimal designs are a well-established tool in adaptive learning \citep{lattimore19bandit} for near-optimal information gathering. Several recent papers applied them to learning reward models in RLHF \citep{das24active,mukherjee24optimal,liu24dual,scheid24optimal}.

We make the following contributions:
\begin{enumerate}
  \item We formalize active learning for DPO as choosing a subset of $n$ data points out of $N$ such the error in DPO logits, the log odds of preferring one response to the other, is minimized (\cref{sec:setting}).
  \item This is the first work that derives a D-optimal design for DPO (\cref{sec:algorithms}). The key idea is to assume log-linear policies, which linearize the DPO objective at the last layer of the neural network policy representation. The derived D-optimal design resembles that of logistic regression, with additional terms due to the reference policy and regularization by it. We propose two computationally-efficient algorithms, \adpo and \adpoplus, which select the most informative data points for DPO. \adpo elicits preferential feedback online and \adpoplus leverages previously logged preferential feedback to have a better design.
  \item We analyze \adpo and \adpoplus, and show that their logit errors are $\tilde{O}(d / \sqrt{n})$, where $d$ is the number of features in the linearized DPO policies and $n$ is the budget on preferential human feedback. This is the first analysis for DPO and has several novel technical aspects. The main technical trick is relating the feedback model and policy parameter under the assumption of log-linear policies. Therefore, we can argue for concentration of the policy parameter with more feedback. The analysis is also under a practical assumption that preferential feedback can be elicited at most once per prompt. To attain a $\tilde{O}(d / \sqrt{n})$ rate in this setting, we introduce a novel assumption on the sufficient diversity of prompts and candidate responses.
  \item We evaluate \adpo and \adpoplus empirically. We experiment with both log-linear DPO policies, which match our theory, and on LLMs. Our methods perform well empirically, despite the fact that they are the first ones with an analysis for active learning in DPO.
\end{enumerate}

The paper is structured as follows. In \cref{sec:background}, we introduce classic methods for training LLMs. In \cref{sec:setting}, we introduce active learning for DPO. We introduce our algorithms in \cref{sec:algorithms} and analyze them in \cref{sec:analysis}. In \cref{sec:experiments}, we evaluate our algorithms empirically. We review related work in detail in \cref{sec:related work} and conclude in \cref{sec:conclusions}.

\section{Background}
\label{sec:background}

We start by introducing our notation. The \emph{prompt} is a string $x \in \cZ$, where $\cZ$ is the space of all strings. The \emph{response} is a string $y \in \cZ$. A \emph{large language model (LLM)} is a \emph{policy} that maps $x$ to $y$. We denote the probability of generating response $y$ to prompt $x$ by a policy parameterized by $\theta \in \Theta$ by $\pi(y \mid x; \theta)$, where $\Theta$ is the space of policy parameters. To simplify terminology, we call $\theta$ a policy when it is clear that we refer to $\pi(\cdot \mid \cdot; \theta)$. Pre-trained LLMs can be optimized by supervised fine-tuning \citep{peft,hu22lora} and reinforcement learning from human feedback, which may require learning of a reward model \citep{ouyang22training} or not \citep{rafailov23direct}. These methods are introduced next.

\subsection{Supervised Fine-Tuning}
\label{sec:sft}

\emph{Supervised fine-tuning (SFT)} \citep{peft,hu22lora} is a direct application of supervised learning to LLMs. The objective of SFT is to minimize the negative \emph{log-likelihood (loglik)} of response $y$ given prompt $x$,
\begin{align}
  \cL_\textsc{sft}(\theta)
  = - \Erv{x, y}{\log \pi(y \mid x; \theta)}\,,
  \label{eq:sft}
\end{align}
in expectation over prompt-response pairs $(x, y)$ sampled from a training set. One limitation of SFT is that we learn only from positive examples. Therefore, it is hard to learn not to generate certain $y$ given $x$. This motivates learning of policies through rewards in \cref{sec:rlhf}.

\subsection{Reinforcement Learning from Human Feedback}
\label{sec:rlhf}

\emph{Reinforcement learning from human feedback (RLHF)} has two stages: reward model learning and policy optimization. The \emph{reward model} $r: \cX \times \cY \to \realset$ is learned from human feedback \citep{ouyang22training}. The LLM policy is then optimized to maximize the expected reward under the reward model using \emph{proximal policy optimization (PPO)} \citep{schulman17proximal}. The objective is
\begin{align}
  & \cL_\textsc{rlhf}(\theta)
  \label{eq:rlhf} \\
  & \!\! = \Erv{x, y \sim \pi(\cdot \mid x; \theta)}{r(x, y) -
  \beta \log \frac{\pi(y \mid x; \theta)}{\pi_0(y \mid x)}}\,,
  \nonumber
\end{align}
where $x$ is a prompt sampled from a training set. The first term is the reward of response $y$ to prompt $x$. The second term penalizes for deviations of policy $\theta$ from a \emph{reference policy} $\pi_0$, usually obtained by SFT (\cref{sec:sft}). The regularization is needed because the reward model is usually learned from data collected by $\pi_0$ and thus cannot estimate the value of significantly different policies well. The parameter $\beta \geq 0$ trades off the two terms. We define the optimal RLHF policy as
$\theta_\textsc{rlhf} = \argmax_{\theta \in \Theta} \cL_\textsc{rlhf}(\theta)$.

\subsection{Direct Preference Optimization}
\label{sec:dpo}

\emph{Direct preference optimization (DPO)} \citep{rafailov23direct} recasts RLHF as follows. Under the \emph{Bradley-Terry-Luce (BTL)} model \citep{bradley52rank,luce05individual} of human feedback, a response with reward $r(x, y_1)$ is preferred to that with reward $r(x, y_2)$ with probability
\begin{align*}
  p(y_1 \succ y_2 \mid x)
  = \mu(r(x, y_1) - r(x, y_2))\,,
\end{align*}
where $\mu(v) = 1 / (1 + \exp[- v])$ is a \emph{sigmoid function}. The key observation in DPO is that the policy that maximizes \eqref{eq:rlhf} has a closed form
\begin{align*}
  \pi(y \mid x; \theta_\textsc{rlhf})
  = \frac{1}{Z(x)} \pi_0(y \mid x) \exp\left[\frac{1}{\beta} r(x, y)\right]\,,
\end{align*}
where $Z(x)$ is the normalizer \citep{rafailov23direct}. This holds for any prompt $x$ and response $y$, under the assumption that the space of optimized policies can represent each conditional distribution exactly. This can be rearranged as $\displaystyle r(x, y) = \beta \log \frac{\pi(y \mid x; \theta_\textsc{rlhf})}{\pi_0(y \mid x)} + \beta Z(x)$ and thus
\begin{align}
  & p(y_1 \succ y_2 \mid x; \theta)
  \label{eq:feedback model} \\
  & \!\! = \mu\left(
  \beta \log \frac{\pi(y_1 \mid x; \theta)}{\pi_0(y_1 \mid x)} -
  \beta \log \frac{\pi(y_2 \mid x; \theta)}{\pi_0(y_2 \mid x)}\right)
  \nonumber
\end{align}
holds when $\theta = \theta_\textsc{rlhf}$. A nice property of this substitution is that the normalizers $Z(x)$, which are difficult to estimate when the space of responses is infinite, cancel out.

Therefore, instead of learning a reward model and optimizing \eqref{eq:rlhf}, we can directly optimize the policy in \eqref{eq:feedback model}. Specifically, let $s \in \set{0, 1}$ be a random variable such that $s = 1$ when $y_1$ is preferred to $y_2$ given $x$, and $s = 0$ when $y_2$ is preferred to $y_1$ given $x$. This problem can be viewed as fitting \eqref{eq:feedback model} to the distribution of $s \mid x, y_1, y_2$ and written as maximizing the negative loglik
\begin{align}
  \cL_\textsc{dpo}(\theta)
  = - \mathbb{E}[&s \log p(y_1 \succ y_2 \mid x; \theta) + {}
  \label{eq:dpo} \\
  & (1 - s) \log p(y_2 \succ y_1 \mid x; \theta)]\,,
  \nonumber
\end{align}
where the expectation is over prompt-candidate response pairs $(x, y_1, y_2)$ sampled from a training set, and stochastic preferential feedback $s \mid x, y_1, y_2$. We define the optimal DPO policy as
\begin{align}
  \textstyle
  \theta_*
  = \argmin_{\theta \in \Theta} \cL_\textsc{dpo}(\theta)
  \label{eq:optimal dpo policy}
\end{align}
and note that it is the \emph{maximum likelihood estimate (MLE)} for \eqref{eq:dpo}. Note that \eqref{eq:dpo} is equivalent to a more classic
\begin{align*}
  \cL_\textsc{dpo}(\theta)
  = - \E{\log p(y_w \succ y_l \mid x; \theta)}
\end{align*}
when the winning response is $y_w = s y_1 + (1 - s) y_2$ and the losing response is $y_l = (1 - s) y_1 + s y_2$. We use the reparameterized objective in \eqref{eq:dpo} because it clearly separates the random variable $s$ from the rest of the objective.

We also note that \eqref{eq:feedback model} can be rewritten as
\begin{align*}
  & p(y_1 \succ y_2 \mid x; \theta) \\
  & \!\! = \mu\left(
  \beta \log \frac{\pi(y_1 \mid x; \theta)}{\pi(y_2 \mid x; \theta)} -
  \beta \frac{\pi_0(y_1 \mid x)}{\pi_0(y_2 \mid x)}\right)\,,
\end{align*}
where $\log \frac{\pi_0(y_1 \mid x)}{\pi_0(y_2 \mid x)}$ depends on the reference policy $\pi_0$ but not on the optimized policy $\theta$. We use this algebraic form because it separates the optimized part of the objective from essentially constants.

\section{Setting}
\label{sec:setting}

We study active learning in DPO (\cref{sec:dpo}). Simply put, instead of assuming that \eqref{eq:dpo} is approximated using a fixed dataset, we choose the dataset actively with the objective of learning policies that are close to $\theta_*$. We study two variants of this problem, offline and online, which we present next.

\textbf{Offline feedback.} The input to this setting is a dataset of size $N$ with preferential human feedback for all data points. The dataset is $\cD = \{(x_i, y_{i, 1}, y_{i, 2}, s_i)\}_{i = 1}^N$, where $x_i$ is the prompt in data point $i \in [N]$, $y_{i, 1}$ and $y_{i, 2}$ are the candidate responses, and $s_i$ is the preferential feedback. Specifically, $s_i = 1$ if the preferred response is $y_{i, 1}$, and $s_i = 0$ if the preferred response is $y_{i, 2}$. Our goal is to select a subset of $\cD$ of size $n$ so that the DPO policy on this subset is \say{close} to $\theta_*$. This setting is motivated by computational efficiency. In particular, even if preferential feedback $s_i$ is known, we may not have computational resources to learn from all of it. Choosing the most informative subset of $\cD$ of size $n$ is a natural way of maximizing the information gain within the computational cost constraint.

\textbf{Online feedback.} The input to this setting is a dataset of size $N$ without preferential human feedback. The dataset is $\cD = \{(x_i, y_{i, 1}, y_{i, 2})\}_{i = 1}^N$, where $x_i$ is the prompt in data point $i \in [N]$, and $y_{i, 1}$ and $y_{i, 2}$ are the candidate responses. The human feedback $s_i$ is elicited online. This setting is motivated by statistical efficiency. We want to collect the most informative feedback using only information about prompts $x_i$, and candidate responses $y_{i, 1}$ and $y_{i, 2}$.

Let $\cS_n \subseteq [N]$ be a subset of $n$ data point indices from $\cD$, either collected online or offline. After the algorithm selects $\cS_n$, we minimize an empirical approximation to \eqref{eq:dpo} on $\cS_n$. Before we define it, we introduce a more compact notation. Let
\begin{align*}
  \mu_i(\theta)
  = \mu\left(\beta \log \frac{\pi(y_{i, 1} \mid x_i; \theta)}
  {\pi(y_{i, 2} \mid x_i; \theta)} - \beta b_i\right)
\end{align*}
be the probability that response $y_{i, 1}$ is preferred to $y_{i, 2}$ given $x_i$ under policy $\theta$, where $b_i = \log\left(\frac{\pi_0(y_{i, 1} \mid x_i)}{\pi_0(y_{i, 2} \mid x_i)}\right)$ is the \emph{bias} due to the reference policy $\pi_0$. Let
\begin{align}
  & \cL_\textsc{dpo}(\theta; \cS)
  \label{eq:negative dpo loglik} \\
  & \!\! = - \sum_{i \in \cS} s_i \log \mu_i(\theta) +
  (1 - s_i) \log(1 - \mu_i(\theta))
  \nonumber
\end{align}
be the \emph{DPO negative loglik} on $\cS \subseteq [N]$. Then \eqref{eq:dpo} can be approximated on $\cS_n$ by $\frac{1}{n} \cL_\textsc{dpo}(\theta; \cS_n)$. We propose algorithms for choosing $\cS_n$ in \cref{sec:algorithms}.

\noindent \textbf{Objective.} Now we are ready to state our objective. Let $\theta_*$ be the optimal DPO policy in \eqref{eq:optimal dpo policy}. Let $\cE(\theta, \theta_*) =$
\begin{align}
  \max_{i \in [N]} \abs{
  \beta \log \frac{\pi(y_{i, 1} \mid x_i; \theta)}
  {\pi(y_{i, 2} \mid x_i; \theta)} -
  \beta \log \frac{\pi(y_{i, 1} \mid x_i; \theta_*)}
  {\pi(y_{i, 2} \mid x_i; \theta_*)}}
  \label{eq:maximum logit error}
\end{align}
be the \emph{maximum logit error} under policy $\theta$, the difference of DPO logits under $\theta$ and $\theta_*$. Note that the biases cancel. Let $\hat{\theta}_n = \argmin_{\theta \in \Theta} \cL_\textsc{dpo}(\theta; \cS_n)$ denote the optimal DPO policy on $\cS_n$. We want $\hat{\theta}_n$ to be close to $\theta_*$ in terms of \eqref{eq:maximum logit error}. Specifically, we want $\cE(\hat{\theta}_n, \theta_*)$ to decrease with $n$ with a high probability. The motivation for \eqref{eq:maximum logit error} is that it can bound many other errors. For instance, since the Lipschitz factor of $\mu$ is $1 / 4$, we get
\begin{align*}
  \max_{i \in [N]} |\mu_i(\hat{\theta}_n) - \mu_i(\theta_*)|
  \leq \frac{1}{4} \cE(\hat{\theta}_n, \theta_*)\,.
\end{align*}
Therefore, when the maximum logit error is small, the estimated probability that $y_{i, 1}$ is preferred to $y_{i, 2}$ under policy $\hat{\theta}_n$, for any data point $i \in [N]$, is close to that under $\theta_*$.

\section{Algorithms}
\label{sec:algorithms}

The key idea in our paper is to linearize the policy at the last layer of its neural network representation and use linear algebra for active learning. Active learning on linearized neural networks was popularized in regret minimization by \citet{riquelme18deep}. \citet{das24active,mukherjee24optimal,thekumparampil24comparing,liu24dual,scheid24optimal} applied it recently to learning reward models. In our work, we linearize policies and formalize it as follows.

\begin{assumption}
\label{ass:log-linear policies} All policies are \emph{log-linear},
\begin{align}
  \pi(y \mid x; \theta)
  \propto \exp[\phi(x, y)\T \theta]\,,
  \label{eq:log-linear policy}
\end{align}
where $\phi(x, y) \in \realset^d$ is the \emph{feature vector} for pair $(x, y)$ and $\theta \in \realset^d$ is a policy parameter.
\end{assumption}

We make this assumption for the rest of the paper. Under this assumption, $\mu_i(\theta)$ in \eqref{eq:negative dpo loglik} becomes
\begin{align}
  \mu_i(\theta)
  = \mu(\beta (\phi_i\T \theta - b_i))\,,
  \label{eq:log-linear probability}
\end{align}
where $\phi_i = \phi(x_i, y_{i, 1}) - \phi(x_i, y_{i, 2})$ is the difference of the feature vectors of responses $y_{i, 1}$ and $y_{i, 2}$ given $x_i$. We note that the normalizers of $\pi(y \mid x; \theta)$ cancel out. We also note that when \eqref{eq:log-linear probability} is substituted into \eqref{eq:negative dpo loglik}, we obtain a similar expression to the negative loglik of logistic regression, except for the bias $b_i$ and $\beta$. The key idea in our algorithms is to optimize the Hessian of the DPO negative loglik.

\begin{lemma}
\label{lem:hessian} Let $\pi(y \mid x; \theta)$ be a log-linear policy. Then the Hessian of $\cL_\textsc{dpo}(\theta; \cS)$ in \eqref{eq:negative dpo loglik} with respect to $\theta$ is
\begin{align*}
  \nabla^2 \cL_\textsc{dpo}(\theta; \cS)
  = \beta^2 \sum_{i \in \cS} \mu_i(\theta) (1 - \mu_i(\theta)) \phi_i \phi_i\T\,.
\end{align*}
It is also positive semi-definite.
\end{lemma}
\begin{proof}
The proof is in \cref{sec:hessian proof}.
\end{proof}

The Hessian $\nabla^2 \cL_\textsc{dpo}(\theta; \cS)$ can be used to derive the covariance matrix of the MLE of $\cL_\textsc{dpo}(\theta; \cS)$ and is also known as the Fisher information matrix \citep{fisher22mathematical}. Therefore, it can be used for both uncertainty quantification and information gathering \citep{lattimore19bandit}. Since the MLE of $\cL_\textsc{dpo}(\theta; \cS)$ is a policy, we can use the Hessian to select a subset of data points to learn better policies.

Specifically, let $\cS_n$ be a subset of $n$ data point indices and $\hat{\theta}_n = \argmin_{\theta \in \Theta} \cL_\textsc{dpo}(\theta; \cS_n)$ be the corresponding MLE. We show in \cref{thm:maximum logit error bound} that the error in the logit estimate at data point $i \in [N]$ is bounded with a high probability as
\begin{align*}
  |\phi_i\T (\hat{\theta}_n - \theta_*)|
  \leq \sqrt{d \phi_i\T (\nabla^2 \cL_\textsc{dpo}(\theta_*; \cS_n))^{-1} \phi_i}
\end{align*}
up to logarithmic factors. To minimize it, we want to maximize all eigenvalues of $\nabla^2 \cL_\textsc{dpo}(\theta_*; \cS_n)$. We achieve this by maximizing $\log\det(\nabla^2 \cL_\textsc{dpo}(\theta_*; \cS_n))$ over $\cS_n$.

This optimization problem is challenging for two reasons. First, it is a discrete optimization problem over $\cS_n$. In our work, we maximize $\log\det(\nabla^2 \cL_\textsc{dpo}(\theta_*; \cS_n))$ greedily. An informal justification for this approach is that $\log\det(X)$ is monotone and concave in $X$ for $X \succeq 0$, and thus a greedy algorithm should be near optimal \citep{nemhauser78approximation}. We prove this formally in \cref{sec:analysis}. Second, $\theta_*$ is unknown. We overcome this by using its plug-in estimates \citep{stufken12optimal}

\subsection{Active DPO with Online Preferential Feedback}
\label{sec:adpo}

\begin{algorithm}[t]
  \caption{\adpo: Active DPO with online feedback.}
  \label{alg:adpo}
  \begin{algorithmic}[1]
    \State \textbf{Input:} Dataset $\cD = \{(x_i, y_{i, 1}, y_{i, 2})\}_{i = 1}^N$
    \State $H_0 \gets \gamma I_d, \ \cS_0 \gets \emptyset$
    \For{$t = 1, \dots, n$}
      \State Solve $\hat{\theta}_{t - 1} \gets \argmin_{\theta \in \Theta}
      \cL_\textsc{dpo}(\theta; \cS_{t - 1})$
      \State Let $v_{t, i} \gets \beta \sqrt{\mu_i(\hat{\theta}_{t - 1})
      (1 - \mu_i(\hat{\theta}_{t - 1}))} \phi_i$
      \State $\displaystyle I_t \gets \argmax_{i \in [N] \setminus \cS_{t - 1}}
      \log\det(H_{t - 1} + v_{t, i} v_{t, i}\T)$
      \State Get preferential feedback $s_{I_t}$ on $(x_{I_t}, y_{I_t, 1}, y_{I_t, 2})$
      \State $H_t \gets H_{t - 1} + v_{t, I_t} v_{t, I_t}$
      \State $\cS_t \gets \cS_{t - 1} + \set{I_t}$
    \EndFor
    \State \textbf{Output:} Data point indices $\cS_n$ for learning a model
  \end{algorithmic}
\end{algorithm}

Our first algorithm does not have access to any preferential feedback initially. It collects it online, re-estimates $\theta_*$, and approximately maximizes $\log\det(\nabla^2 \cL_\textsc{dpo}(\theta_*; \cS_n))$.

The pseudo-code of the algorithm is in \cref{alg:adpo} and we call it \emph{active DPO (\adpo)}. \adpo chooses data points in $n$ rounds. The indices of the chosen data points in the first $t$ rounds are denoted by $S_t$ and the corresponding Hessian is $H_t$. We refer to it as the \emph{design matrix} since it is used to select next data points. The design matrix is initialized to $\gamma I_d$, where $\gamma > 0$ is a constant that guarantees that all $H_t$ are well defined. In round $t$, \adpo selects the index $I_t$ that greedily maximizes the information gain given $H_t$ and the empirical estimate of $\theta_*$ up to round $t$, $\hat{\theta}_{t - 1}$ (line 6). This is because
\begin{align*}
  v_{t, i} v_{t, i}\T
  = \beta^2 \mu_i(\hat{\theta}_{t - 1}) (1 - \mu_i(\hat{\theta}_{t - 1}))
  \phi_i \phi_t\T
\end{align*}
can be viewed as the incremental gain due to data point $i$ in \cref{lem:hessian}. After the data point $I_t$ is chosen, we observe preferential feedback on it (line 7) and update all statistics (lines 8-9). Finally, after $n$ rounds, \adpo outputs $n$ chosen indices (line 10) and an LLM policy is optimized on them using DPO.

The time complexity of \adpo is $O(n^2 + n N)$. The former term is due to training on all past feedback in each round (line 4) and the latter is due to maximizing exactly in line 6. In experiments, we reduce the former to $O(n \log n)$ by estimating $\hat{\theta}_{t - 1}$ only a logarithmic number of times, when $t = 2^i$ for some integer $i > 0$. We reduce the latter to $O(n)$ by replacing $[N] \setminus \cS_{t - 1}$ with its random subset of a fixed size $256$. Finally, note that $I_t$ in line $6$ can be equivalently expressed (\cref{sec:greedy logdet maximization proof}) as
\begin{align}
  \textstyle
  I_t
  = \argmax_{i \in [N] \setminus \cS_{t - 1}} v_{t, i} H_{t - 1}^{-1} v_{t, i}\T\,.
  \label{eq:acquisition function}
\end{align}
Therefore, the determinant does not need to be computed. The inverse $H_{t - 1}^{-1}$ can be computed incrementally using the Sherman-Morrison formula, with $O(d^2)$ update time. The statistical efficiency of \adpo is analyzed in \cref{sec:analysis}.

\subsection{Active DPO with Offline Preferential Feedback}
\label{sec:adpo+}

Our second algorithm has access to preferential feedback initially. All feedback is used to estimate $\theta_*$, which is then used to approximately maximize $\log\det(\nabla^2 \cL_\textsc{dpo}(\theta_*; \cS_n))$.

The pseudo-code of our algorithm is in \cref{alg:adpo+} and we call it \adpoplus, where $+$ indicates that \adpoplus has access to more information than \adpo. \adpoplus differs from \adpo in two steps. First, $\theta_*$ is estimated initially (line 3) from all preferential feedback. Second, no preferential feedback is collected online. Similarly to \adpo, the time complexity of \adpoplus is $O(n N)$ because of the exact maximization in line 7. We reduce it to $O(n)$ in experiments as in \cref{sec:adpo}. 

\begin{algorithm}[t]
  \caption{\adpoplus: Active DPO for offline feedback.}
  \label{alg:adpo+}
  \begin{algorithmic}[1]
    \State \textbf{Input:} Dataset $\cD = \{(x_i, y_{i, 1}, y_{i, 2}, s_i)\}_{i = 1}^N$
    \State $H_0 \gets \gamma I_d, \ \cS_0 \gets \emptyset$
    \State Solve $\hat{\theta} \gets \argmin_{\theta \in \Theta}
    \cL_\textsc{dpo}(\theta; [N])$
    \For{$t = 1, \dots, n$}
      \State $\hat{\theta}_{t - 1} \gets \hat{\theta}$
      \State Let $v_{t, i} \gets \beta \sqrt{\mu_i(\hat{\theta}_{t - 1})
      (1 - \mu_i(\hat{\theta}_{t - 1}))} \phi_i$
      \State $\displaystyle I_t \gets \argmax_{i \in [N] \setminus \cS_{t - 1}}
      \log\det(H_{t - 1} + v_{t, i} v_{t, i}\T)$
      \State $H_t \gets H_{t - 1} + v_{t, I_t} v_{t, I_t}$
      \State $\cS_t \gets \cS_{t - 1} + \set{I_t}$
    \EndFor
    \State \textbf{Output:} Data point indices $\cS_n$ for learning a model
  \end{algorithmic}
\end{algorithm}

\section{Analysis}
\label{sec:analysis}

In this section, we provide a unified analysis for \adpo and \adpoplus. This is possible because the algorithms only differ in how the instance-specific factors in the design matrix are estimated. In \adpoplus, they are estimated from all preferential feedback. In \adpo, only the online elicited feedback up to round $t$ is used. We state our assumptions first.

We assume that all policies are log-linear (\cref{ass:log-linear policies}) and that the collected feedback $s_{I_t}$ is conditionally independent given all feedback up to round $t$, for all $t \in [n]$. Under this assumption, the negative loglik in \eqref{eq:negative dpo loglik} is similar to that of logistic regression and we can use existing concentration inequalities \citep{abbasi-yadkori11improved}.

\begin{assumption}
\label{ass:boundedness}[Boundedness] For any $i \in [N]$, $\normw{\phi_i}{2} \leq 1$ and $|b_i| \leq 1$. We assume that $\Theta$ is a unit sphere, and hence $\normw{\theta_*}{2} \leq 1$ and $\normw{\hat{\theta}_n}{2} \leq 1$.
\end{assumption}

Assumptions on feature vectors, comprising $\phi_i$ and $b_i$, are standard in the analyses of generalized linear models \citep{li17provably,kveton20randomized,mukherjee24optimal}. Our assumption on $\theta_*$ and $\hat{\theta}_n$ can be guaranteed by applying DPO to a unit sphere $\Theta$. The assumption can be weakened to $\normw{\hat{\theta}_n - \theta_*}{2} \leq 1$ using initial exploration \citep{li17provably,kveton20randomized}.

We can analyze \adpo and \adpoplus in a unified way because the instance-specific factors in their design matrices can be bounded from below by $c_{\min}$ and above by $c_{\max}$.

\begin{assumption}
\label{ass:design matrix}[Design matrix] For any $i \in [N]$ and $\theta \in \Theta$, we have $0 \leq c_{\min} \leq \beta^2 \mu_i(\theta) (1 - \mu_i(\theta)) \leq c_{\max}$.
\end{assumption}

These constants obviously exist and can be easily derived. For instance, since $\max_{x \in \realset} \mu(x) (1 - \mu(x)) = 0.25$, we get $c_{\max} = 0.25 \beta^2$. Moreover, under \cref{ass:boundedness}, we have for any $\mu_i(\theta) \leq 0.5$ that
\begin{align*}
  \beta^2 \mu_i(\theta) (1 - \mu_i(\theta))
  \geq \beta^2 \mu_i^2(\theta)
  \geq \beta^2 \mu(- 4 \beta)
  = c_{\min}\,.
\end{align*}
The argument for $\mu_i(\theta) \geq 0.5$ is similar. The constants $c_{\min}$ and $c_{\max}$ appear in our bounds.

The last assumption is that the dataset is sufficiently diverse.

\begin{assumption}
\label{ass:diverse dataset}[Diverse dataset] There exists a constant $\kappa \geq 1$ such that $v_{t, i}\T H_{t - 1}^{-1} v_{t, i} \leq \kappa v_{t, I_t}\T H_{t - 1}^{-1} v_{t, I_t}$ holds for any $i \in [N]$ and $t \in [n]$.
\end{assumption}

This assumption says that the maximizer in \eqref{eq:acquisition function} is an approximate upper bound, up to a multiplicative $\kappa \geq 1$, on the information gain at each data point, including those previously chosen that cannot be chosen again. We note that the assumption holds for $\kappa = 1$ when repeated independent observations of the data points are allowed, as in all prior works (\cref{sec:related work}). In this case, the maximization in \eqref{eq:acquisition function} would be over $i \in [N]$.

\subsection{Main Result}
\label{sec:main result}

We state our main claim below.

\begin{theorem}
\label{thm:maximum logit error bound} Let $\hat{\theta}_n = \argmin_{\theta \in \Theta} \cL_\textsc{dpo}(\theta; \cS_n)$. Then the maximum logit error under \adpo and \adpoplus is
\begin{align*}
  \cE(\hat{\theta}_n, \theta_*)
  = \tilde{O}(d \sqrt{\log(1 / \delta) / n})
\end{align*}
with probability at least $1 - \delta$, where $\tilde{O}$ hides all logarithmic factors but those in $\delta$.
\end{theorem}

We prove the claim as follows. For log-linear policies, \eqref{eq:maximum logit error} reduces to $\max_{i \in [N]} |\phi_i\T (\hat{\theta}_n - \theta_*)|$. By the Cauchy-Schwarz inequality, for any data point $i \in [N]$,
\begin{align}
  |\phi_i\T (\hat{\theta}_n - \theta_*)|
  \leq \normw{\phi_i}{\Sigma_n^{-1}}
  \normw{\hat{\theta}_n - \theta_*}{\Sigma_n}\,,
  \label{eq:logit error bound}
\end{align}
where $\Sigma_n = \gamma I_d + \nabla^2 \cL_\textsc{dpo}(\theta_*; \cS_n)$ a regularized Hessian at the optimal DPO policy $\theta_*$. To bound the first term, we note that the feedback at data point $i$ is distributed as
\begin{align}
  s_i
  \sim \mu_i(\theta_*)
  = \mu(\beta (\phi_i\T \theta_* - b_i))\,.
  \label{eq:independent feedback}
\end{align}
This assumption follows from the definition of DPO in \eqref{eq:feedback model}, which says that $\mu_i(\theta_*)$ is the probability that response $y_{i, 1}$ is preferred to $y_{i, 2}$ given $x_i$. Thus we can build on existing concentration results for sub-Gaussian random variables to prove the following.

\begin{theorem}
\label{thm:self-normalization} For any set of $n$ indices $\cS_n \subseteq [N]$,
\begin{align*}
  \normw{\hat{\theta}_n - \theta_*}{\Sigma_n}
  \leq \sqrt{\frac{\beta^2 d}{c_{\min}}
  \log\left(\frac{1 + c_{\min} n / \gamma}{\delta}\right)} +
  2 \gamma^{\frac{1}{2}}
\end{align*}
holds with probability at least $1 - \delta$.
\end{theorem}

To bound the second term in \eqref{eq:logit error bound}, we use the fact that the standard errors of the logit estimates do not increase over time and decrease at a desired rate if \cref{ass:diverse dataset} holds for some constant $\kappa \geq 1$.

\begin{theorem}
\label{thm:greedy logdet maximization} For any data point $i \in [N]$,
\begin{align*}
  \phi_i\T \Sigma_n^{-1} \phi_i
  \leq \frac{c_{\max}^3
  \log\left(1 + \frac{c_{\max} n}{\gamma d}\right)}
  {c_{\min} \gamma \log(1 + c_{\max} / \gamma)} \frac{\kappa d}{n}\,.
\end{align*}
\end{theorem}

All proofs are in \cref{sec:proofs}.

\subsection{Discussion}
\label{sec:discussion}

The bound in \cref{thm:maximum logit error bound} is $\tilde{O}(d \sqrt{\log(1 / \delta) / n})$ and holds with probability at least $1 - \delta$. As a result, the maximum logit error decreases with more feedback $n$ and increases with the number of learned policy parameters $d$. The bound is not directly comparable to prior works in \cref{sec:related work} because they bound reward model errors, while we bound a policy learning error. That being said, the dependence on $n$ and $\delta$ is similar. The linear dependence on $d$ arises because \cref{thm:greedy logdet maximization} is proved through a self-normalizing bound in \cref{thm:self-normalization} that would apply even to infinitely-large datasets. We would get an $\tilde{O}(\sqrt{d \log(N) \log(1 / \delta) / n})$ bound, where $N$ is the dataset size, if we followed the analysis of \citet{kveton20randomized} and applied a union bound over all data points.

\section{Experiments}
\label{sec:experiments}

\begin{figure*}[t!]
  \centering
  \includegraphics[width=6.4in]{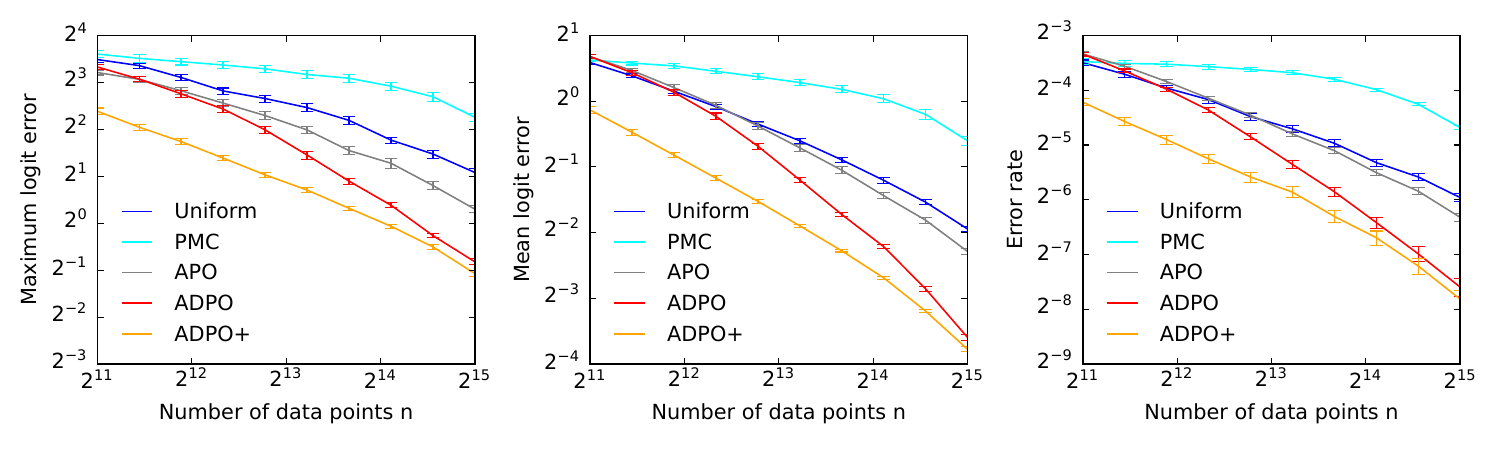}
  \vspace{-0.1in} \\
  \includegraphics[width=6.4in]{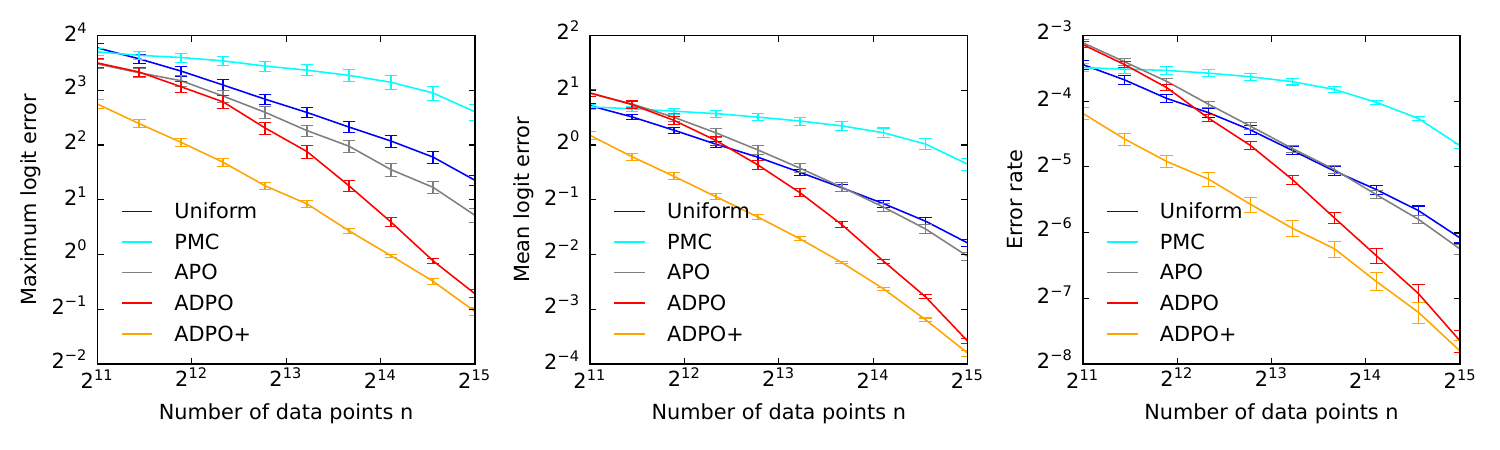}
  \vspace{-0.15in}
  \caption{Experiments with log-linear policies on the CIFAR-10 (first row) and CIFAR-100 (second row) datasets.}
  \label{fig:cifar}
\end{figure*}

We experiment with both log-linear (\cref{sec:log-linear policies}) and LLM (\cref{sec:llm policies}) policies. The log-linear experiments validate that \adpo and \adpoplus work as analyzed. The LLM experiments show that \adpo and \adpoplus perform well in practice when applied to LLMs. We conduct more experiments with log-linear policies in \cref{sec:ablation study}.

\subsection{Log-Linear Policies}
\label{sec:log-linear policies}

This experiment is designed as follows. First, we take an existing multi-class classification dataset and turn it into a preferential feedback dataset. More specifically, we choose a random positive label and generate $N$ vectors $\set{\phi_i}_{i = 1}^N$, where $\phi_i \in \realset^d$ is the difference of feature vectors of random positive and negative examples. Second, we label all $\phi_i$ with $1$ and learn a logistic regression model to simulate preferential feedback. Let $\bar{\theta}$ and $\bar{\Sigma}$ be the learned model parameter and its covariance, respectively. Third, we generate preferential feedback $s_i \sim \mathrm{Ber}(\mu(\phi_i\T \bar{\theta}))$ for all $\phi_i$ and get a dataset $\cD = \set{(\phi_i, s_i)}_{i = 1}^N$. Fourth, we generate a reference policy as $\theta_0 \sim \cN(\bar{\theta}, \bar{\Sigma})$ and set the bias as $b_i = \phi_i\T \theta_0$. Simply put, $\theta_0$ is close $\bar{\theta}$, as measured by the uncertainty of $\bar{\theta}$. Finally, we compute the optimal DPO policy $\theta_*$ on $\cD$. All compared methods apply DPO to their selected subset $\cS_n$ of $\cD$ and learn $\hat{\theta}_n = \argmin_{\theta \in \Theta} \cL_\textsc{dpo}(\theta; \cS_n)$.

We compare $\hat{\theta}_n$ to $\theta_*$ in three metrics. The first metric is the \emph{maximum logit error}, $\max_{i \in [N]} |\phi_i\T (\hat{\theta}_n - \theta_*)|$, which we bound in \cref{thm:maximum logit error bound}. The second metric is the \emph{mean logit error} $\frac{1}{N} \sum_{i = 1}^N |\phi_i\T (\hat{\theta}_n - \theta_*)|$. Although we do not analyze it, our methods minimize it indirectly through the maximum error. The last metric is the \emph{error rate},
\begin{align*}
  \frac{1}{N} \sum_{i = 1}^N \I{\mathrm{sgn}(\phi_i\T \hat{\theta}_n - b_i) \neq
  \mathrm{sgn}(\phi_i\T \theta_* - b_i)}\,,
\end{align*}
which is the fraction of incorrectly ordered responses by $\hat{\theta}_n$ when $\theta_*$ is the ground truth.

\begin{figure*}[t!]
  \centering
  \includegraphics[width=6.4in]{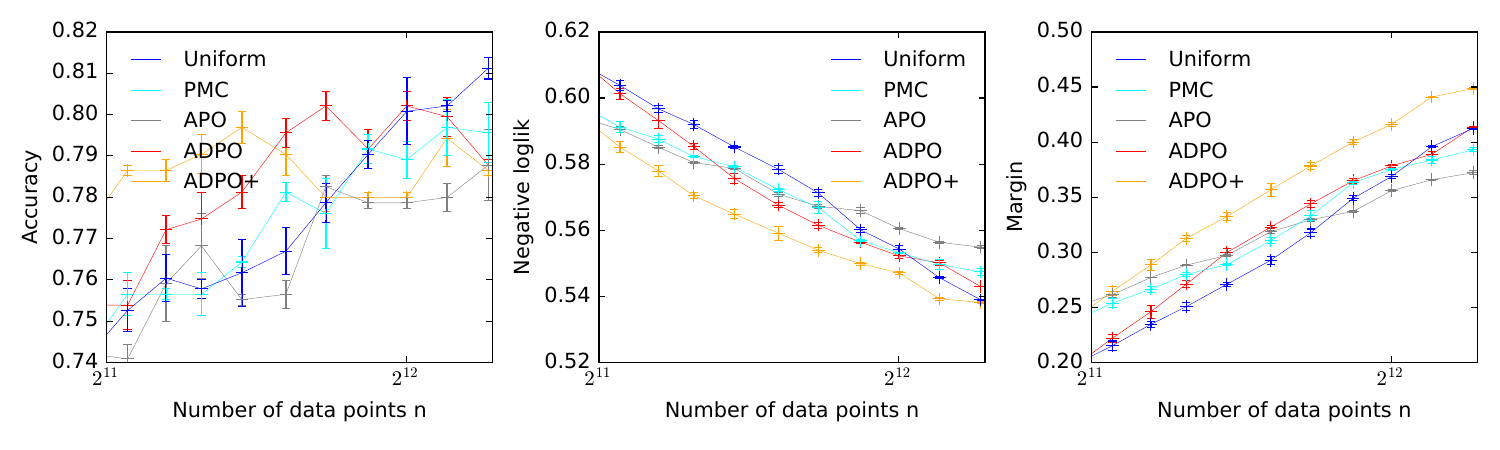}
  \vspace{-0.1in} \\
  \includegraphics[width=6.4in]{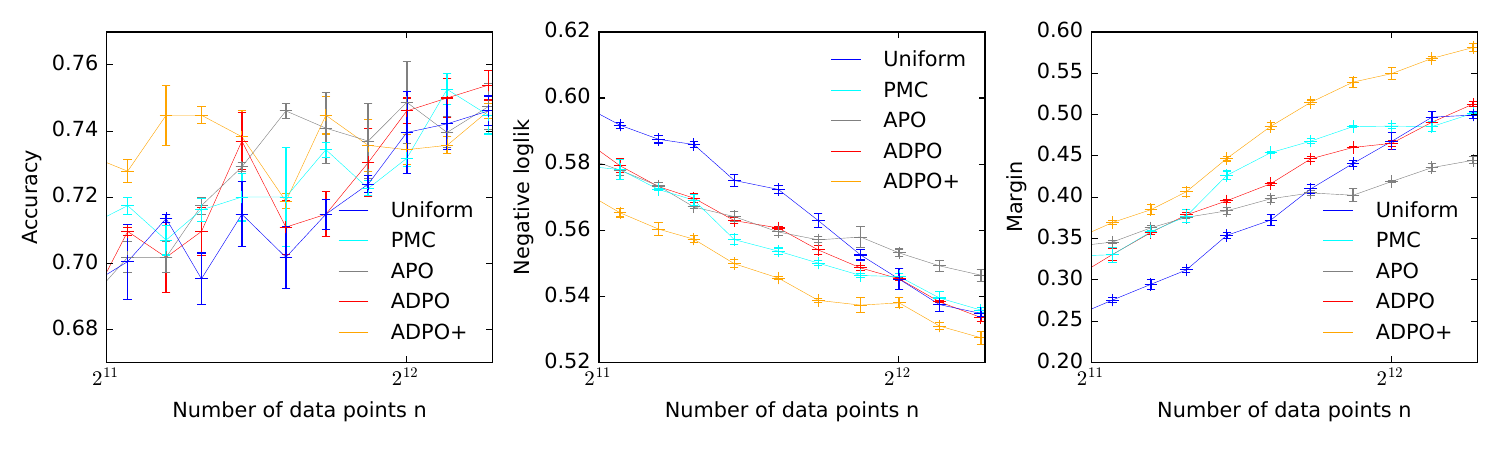}
  \vspace{-0.15in}
  \caption{Experiments with LLM policies on the Nectar dataset. We use Llama-3.2 (first row) and Phi-3 (second row) models.}
  \label{fig:llms}
\end{figure*}

We compare five algorithms. The first two algorithms are \adpo and \adpoplus. We expect \adpoplus to perform better because it has access to more information. We consider three baselines: \uniform, \apo, and \pmc. \uniform selects data points uniformly at random. While simple, it is known to be competitive in real-world problems where feature vectors may cover the feature space close to uniformly \citep{ash20deep,ash21gone,mukherjee24optimal,muldrew2024active}. \apo is the practical incremental D-optimal design for linear models proposed in \citet{das24active}. The main difference from \adpo is that \apo neglects logistic model factors and $\beta$ (\cref{lem:hessian}). Therefore, while it selects diverse $\phi_i$, they do not necessarily maximize the information gain in DPO. The last baseline is \pmc of \citet{muldrew2024active}, which selects data points with the highest differences between estimated rewards of their responses.

We experiment with CIFAR-10 and CIFAR-100 datasets \citep{kriz09learning}. The features are a random subset of ResNet-50 embeddings \citep{he16deep} of size $d = 384$. The dataset size is $N = 2^{16}$. We set the DPO regularizer to $\beta = 1$ and experiment with other $\beta$ in \cref{sec:ablation study}. Our CIFAR-10 results are reported in the first row of \cref{fig:cifar}. \adpoplus is the best performing method in all metrics. Many improvements are major. For instance, the lowest maximum logit error of \uniform ($n = 2^{15}$) is attained by \adpoplus at $n < 2^{13}$. The lowest maximum logit error of \apo ($n = 2^{15}$) is attained by \adpoplus at $n < 2^{14}$. \adpo is the second best method in the maximum logit error. It is never worse than \uniform, \apo, and \pmc. \adpo improves in all metrics over all baselines at larger sample sizes. Our CIFAR-100 results are reported in the second row of \cref{fig:cifar} and we observe the same trends as on the CIFAR-10 dataset.

\subsection{LLM Policies}
\label{sec:llm policies}

We also experiment with a real-world preference dataset Nectar \cite{starling2023} and two LLM policies: Llama-3.2 (3B parameters) \cite{dubey2024llama} and Phi-3 \cite{abdin2024phi}.
We sample $N = 5\,000$ prompts $\set{x_i}_{i = 1}^N$ from the dataset, each with two responses. The accepted $\set{y_{i, w}}_{i = 1}^N$ and rejected $\set{y_{i, l}}_{i = 1}^N$ responses are determined based on the ground truth in the dataset. The feature vector $\phi(x, y)$ is the embedding of the concatenated prompt and response from the last hidden layer of the LLM, of size $d = 4\,096$. The bias term is $b_i = \log \pi_0(y_{i, w} \mid x_i) - \log \pi_0(y_{i, l} \mid x_i)$, where $\pi_0$ is the initial LLM reference policy.

We report three metrics. The \emph{accuracy} measures how well we distinguish between positive and negative responses,
\begin{align*}
  \frac{1}{N} \sum_{i = 1}^N \I{
  \log \frac{\pi(y_{i, w} \mid x_i; \theta)}{\pi_0(y_{i, w} \mid x_i)} >
  \log \frac{\pi(y_{i, l} \mid x_i; \theta)}{\pi_0(y_{i,l} \mid x_i)}}\,.
\end{align*}
This metric is $1$ minus the error rate in \cref{fig:cifar} and thus identical, up to how we plot it. We could not plot the two other metrics in \cref{fig:cifar} because they require knowing $\theta_*$. Therefore, we decided to plot two other metrics that reflect the confidence in distinguishing the responses. The \emph{margin} is the advantage of a positive response over a negative one,
\begin{align*}
  \frac{1}{N} \sum_{i = 1}^N
  \beta \log \frac{\pi(y_{i, w} \mid x_i; \theta)}{\pi_0(y_{i, w} \mid x_i)} -
  \beta \log \frac{\pi(y_{i, l} \mid x_i; \theta)}{\pi_0(y_{i, l} \mid x_i)}\,.
\end{align*}
The \emph{negative loglik} is the logistic regression loss,
\begin{align*}
  - \frac{1}{N} \sum_{i = 1}^N \log \mu\left(
  \beta \log \frac{\pi(y_{i, w} \mid x_i; \theta)}{\pi_0(y_{i, w} \mid x_i)} -
  \beta \log \frac{\pi(y_{i, l} \mid x_i; \theta)}{\pi_0(y_{i, l} \mid x_i)}\right)\,.
\end{align*}
Our results with Llama-3.2 and Phi-3 models are reported in \cref{fig:llms}. We observe similar trends to \cref{fig:cifar}. \adpoplus is clearly the best performing method in both the margin and negative loglik. \adpo is among the best three methods for larger sample sizes. The least clear trend is in accuracy. We believe that this is because many responses are of a similar quality. Therefore, they cannot be easily distinguished and lie close to the decision boundary, which can be impacted by even minor changes in the LLM.

\section{Conclusions}
\label{sec:conclusions}

We propose an active learning framework for DPO. The key idea is to linearize the DPO objective at the last layer of the neural network representation of the optimized policy and then compute the D-optimal design to collect preferential feedback. We propose two algorithms. One is for the online setting, where the human feedback is elicited online, and the other is for the offline setting, where the feedback has already been collected and we choose its subset to improve the computation efficiency of DPO. We analyze both algorithms and also evaluate them empirically, in the setting that matches our theory and on LLMs.

This is the first work that applies optimal designs to DPO. The main difference from prior works is that the optimal design is applied to policy optimization. A natural direction for future work are other policy optimization frameworks, such as KTO \citep{ethayarajh24model}. Our analysis could also be improved in several aspects. For instance, it is for log-linear policies and we have not derived an upper bound on $\kappa$ in \cref{ass:diverse dataset}. In the setting of prior works, where multiple independent observations of preferential feedback for the same prompt are possible, $\kappa = 1$.



\bibliographystyle{plainnat}
\bibliography{Brano,Ryan,Tong}

\clearpage
\onecolumn
\appendix

\section{Proofs and Supporting Lemmas}
\label{sec:proofs}

This section contains proofs of our main claims and supporting lemmas.

\subsection{Proof of \cref{lem:hessian}}
\label{sec:hessian proof}

Let $v \in \realset$ and $\mu(v) = 1 / (1 + \exp[- v])$. Then
\begin{align*}
  \frac{\partial}{\partial v} \mu(v)
  = - \frac{1}{(1 + \exp[- v])^2} \frac{\partial}{\partial v} \exp[- v]
  = \frac{\exp[- v]}{(1 + \exp[- v])^2}
  = \mu(v) (1 - \mu(v))\,.
\end{align*}
We start with computing the gradient of \eqref{eq:negative dpo loglik},
\begin{align*}
  \nabla \cL_\textsc{dpo}(\theta; \cS)
  & = - \sum_{i \in \cS} s_i \frac{\nabla \mu_i(\theta)}{\mu_i(\theta)} -
  (1 - s_i) \frac{\nabla \mu_i(\theta)}{1 - \mu_i(\theta)}
  = \beta \sum_{i \in \cS} (1 - s_i) \mu_i(\theta) \phi_i -
  s_i (1 - \mu_i(\theta)) \phi_i \\
  & = \beta \sum_{i \in \cS} (\mu_i(\theta) - s_i) \phi_i\,.
\end{align*}
It follows that the Hessian is
\begin{align*}
  \nabla^2 \cL_\textsc{dpo}(\theta; \cS)
  = \nabla (\nabla \cL_\textsc{dpo}(\theta; \cS))
  = \beta \sum_{i \in \cS} \phi_i \nabla \mu_i(\theta)
  = \beta^2 \sum_{i \in \cS} \mu_i(\theta) (1 - \mu_i(\theta)) \phi_i \phi_i\T\,.
\end{align*}
The term $\phi_i \phi_i\T$ is an outer product, which is positive semi-definite. Because $\mu_i(\theta) (1 - \mu_i(\theta)) \geq 0$, the Hessian is a weighted sum of positive semi-definite matrices, and thus a positive semi-definite matrix.

\subsection{Proof of \cref{thm:self-normalization}}
\label{sec:self-normalization proof}

Let $\hat{\Sigma}_n = \nabla^2 \cL_\textsc{dpo}(\theta_*; \cS_n)$. We start by noting that $\hat{\Sigma}_n$ is a positive semi-definite matrix (\cref{lem:hessian}). Therefore, $\cL_\textsc{dpo}(\theta; \cS_n)$ is strongly convex in $\theta$ and
\begin{align*}
  \cL_\textsc{dpo}(\hat{\theta}_n; \cS_n)
  & \geq \cL_\textsc{dpo}(\theta_*; \cS_n) +
  \langle\nabla \cL_\textsc{dpo}(\theta_*; \cS_n), \hat{\theta}_n - \theta_*\rangle +
  \frac{1}{2} \normw{\hat{\theta}_n - \theta_*}{\hat{\Sigma}_n}^2
\end{align*}
holds. Now we use that $\cL_\textsc{dpo}(\theta_*; \cS_n) \geq \cL_\textsc{dpo}(\hat{\theta}_n; \cS_n)$ and that $\hat{\Sigma}_n = \Sigma_n - \gamma I_d$, rearrange the inequality, and get
\begin{align*}
  \normw{\hat{\theta}_n - \theta_*}{\Sigma_n}^2
  \leq 2 \langle\nabla \cL_\textsc{dpo}(\theta_*; \cS_n),
  \theta_* - \hat{\theta}_n\rangle +
  \gamma \normw{\hat{\theta}_n - \theta_*}{2}^2\,.
\end{align*}
Then we apply the Cauchy–Schwarz inequality to the right-hand side and get
\begin{align*}
  \normw{\hat{\theta}_n - \theta_*}{\Sigma_n}^2
  \leq 2 \normw{\nabla \cL_\textsc{dpo}(\theta_*; \cS_n)}{\Sigma_n^{-1}}
  \normw{\hat{\theta}_n - \theta_*}{\Sigma_n} +
  \gamma \normw{\hat{\theta}_n - \theta_*}{2}^2\,.
\end{align*}
Now we divide both sides by $\normw{\hat{\theta}_n - \theta_*}{\Sigma_n} > 0$ and get
\begin{align*}
  \normw{\hat{\theta}_n - \theta_*}{\Sigma_n}
  \leq 2 \normw{\nabla \cL_\textsc{dpo}(\theta_*; \cS_n)}{\Sigma_n^{-1}} +
  \frac{\gamma \normw{\hat{\theta}_n - \theta_*}{2}^2}
  {\normw{\hat{\theta}_n - \theta_*}{\Sigma_n}}
  \leq 2 \normw{\nabla \cL_\textsc{dpo}(\theta_*; \cS_n)}{\Sigma_n^{-1}} +
  2 \gamma^{\frac{1}{2}}\,.
\end{align*}
The last inequality follows from
\begin{align*}
  \normw{\hat{\theta}_n - \theta_*}{\Sigma_n}
  = \sqrt{(\hat{\theta}_n - \theta_*)\T \Sigma_n (\hat{\theta}_n - \theta_*)}
  \geq \sqrt{\gamma} \normw{\hat{\theta}_n - \theta_*}{2}^2\,,
\end{align*}
which is proved using $\Sigma_n \succeq \gamma I_d$, and that $\normw{\hat{\theta}_n - \theta_*}{2} \leq 2$.

Therefore, to bound $\normw{\hat{\theta}_n - \theta_*}{\Sigma_n}$, it suffices to show that $\normw{\nabla \cL_\textsc{dpo}(\theta_*; \cS_n)}{\Sigma_n^{-1}}$ is small with a high probability. We show this next. We start by recalling from \cref{lem:hessian} that
\begin{align*}
  \nabla \cL_\textsc{dpo}(\theta_*; \cS_n)
  = \beta \sum_{i \in \cS_n} (\mu_i(\theta_*) - s_i) \phi_i\,,
\end{align*}
where $s_i$ is a binary random variable with mean $\E{s_i} = \mu_i(\theta_*)$, as described in (\ref{eq:independent feedback}). Let $Z_i = \mu_i(\theta_*) - s_i$. Since
\begin{align*}
  \Sigma_n
  \succeq c_{\min} \left(\frac{\gamma}{c_{\min}} I_d +
  \sum_{i \in \cS_n} \phi_i \phi_i\T\right)\,,
\end{align*}
we get
\begin{align*}
  \normw{\nabla \cL_\textsc{dpo}(\theta_*; \cS_n)}{\Sigma_n^{-1}}
  \leq \frac{\beta}{\sqrt{c_{\min}}}
  \Big\|\sum_{i \in \cS_n} Z_i \phi_i\Big\|_{V_n^{-1}}
\end{align*}
for $V_n = \gamma I_d / c_{\min} + \sum_{i \in \cS_n} \phi_i \phi_i\T$. Finally, since $s_i$ are conditionally independent given the history and their variance proxy is $0.25$, we can use Theorem 1 of \citet{abbasi-yadkori11improved} and get that
\begin{align*}
  \Big\|\sum_{i \in \cS_n} Z_i \phi_i\Big\|_{V_n^{-1}}
  \leq \sqrt{\frac{d}{4} \log\left(\frac{1 + c_{\min} n / \gamma}{\delta}\right)}
\end{align*}
holds with probability at least $1 - \delta$. Finally, we collect all inequalities and get that
\begin{align*}
  \normw{\hat{\theta}_n - \theta_*}{\Sigma_n}
  \leq \normw{\nabla \cL_\textsc{dpo}(\theta_*; \cS_n)}{\Sigma_n^{-1}} +
  2 \gamma^{\frac{1}{2}}
  \leq \sqrt{\frac{\beta^2 d}{c_{\min}}
  \log\left(\frac{1 + c_{\min} n / \gamma}{\delta}\right)} +
  2 \gamma^{\frac{1}{2}}
\end{align*}
holds with probability at least $1 - \delta$.

\subsection{Proof of \cref{thm:greedy logdet maximization}}
\label{sec:greedy logdet maximization proof}

First, we introduce $\mu_{t, i} = \mu_i(\hat{\theta}_{t - 1})$, and note that $v_{t, i}$ in \adpo and \adpoplus can be redefined as
\begin{align*}
  v_{t, i}
  = \beta \sqrt{\mu_{t, i} (1 - \mu_{t, i})} \phi_i\,.
\end{align*}
Now note that
\begin{align*}
  \normw{\phi_i}{\Sigma_n^{-1}}^2
  = \phi_i\T \Sigma_n^{-1} \phi_i
  \leq \frac{c_{\max}}{c_{\min}} \phi_i\T H_n^{-1} \phi_i
\end{align*}
because $H_t = \gamma I_d + \sum_{i \in \cS_t} v_{t, i} v_{t, i}\T$. Next we utilize the fact that the standard errors of the estimates decrease with more observations.

\begin{lemma}
\label{lem:variance monotonicity} For any $i \in [N]$ and $t \in [n]$,
\begin{align*}
  \phi_i\T H_t^{-1} \phi_i
  \leq \phi_i\T H_{t - 1}^{-1} \phi_i\,.
\end{align*}
\end{lemma}
\begin{proof}
The proof follows from the Sherman–Morrison formula. Specifically, since
\begin{align*}
  H_t^{-1}
  = H_{t - 1}^{-1} -
  \frac{H_{t - 1}^{-1} \phi_i \phi_i\T H_{t - 1}^{-1}}
  {1 + \phi_i\T H_{t - 1}^{-1} \phi_i}
  \preceq H_{t - 1}^{-1}\,,
\end{align*}
we get $v\T H_t^{-1} v \leq v\T H_{t - 1}^{-1} v$ for any vector $v \in \realset^d$. This completes the proof.
\end{proof}

\cref{lem:variance monotonicity} implies that
\begin{align*}
  \phi_i\T H_n^{-1} \phi_i
  \leq \frac{1}{n} \sum_{t = 1}^n \phi_i\T H_{t - 1}^{-1} \phi_i
  \leq \frac{c_{\max}}{n} \sum_{t = 1}^n v_{t, i}\T H_{t - 1}^{-1} v_{t, i}
\end{align*}
holds for any $i \in [N]$. This allows us to attribute the quality of the solution to individual greedy steps in \adpo and \adpoplus. The next step is to relate $v_{t, i}\T H_{t - 1}^{-1} v_{t, i}$ to $v_{t, I_t}\T H_{t - 1}^{-1} v_{t, I_t}$. The key observation is that
\begin{align*}
  I_t
  & = \argmax_{i \in [N] \setminus \cS_{t - 1}}
  \log\det(H_{t - 1} + v_{t, i} v_{t, i}\T)
  = \argmax_{i \in [N] \setminus \cS_{t - 1}} \log\det(I_d +
  H_{t - 1}^{- \frac{1}{2}} v_{t, i}
  v_{t, i}\T H_{t - 1}^{- \frac{1}{2}}) \\
  & = \argmax_{i \in [N] \setminus \cS_{t - 1}}
  \log(1 + v_{t, i}\T H_{t - 1}^{-1} v_{t, i})
  = \argmax_{i \in [N] \setminus \cS_{t - 1}}
  v_{t, i}\T H_{t - 1}^{-1} v_{t, i}\,.
\end{align*}
The second equality holds because $H_{t - 1}$ is fixed when $I_t$ is selected. The last equality holds because the logarithm is a monotone function. It follows that $I_t$ is the index of the feature vector with the maximum variance.

If the scope of the maximization was $i \in [N]$, the inequality $v_{t, i}\T H_{t - 1}^{-1} v_{t, i} \leq v_{t, I_t}\T H_{t - 1}^{-1} v_{t, I_t}$ would hold for any $i \in [N]$. Since the scope is $i \in [N] \setminus \cS_{t - 1}$, we make \cref{ass:diverse dataset}, which equates to assuming that $\phi_i$ are sufficiently diverse. We also use the following logarithmic transformation.

\begin{lemma}
\label{lem:logarithmic transformation} For any $v \in \realset^d$ and $t \in [n]$,
\begin{align*}
  v\T H_{t - 1}^{-1} v
  \leq \frac{c_{\max}}{\gamma \log(1 + c_{\max} / \gamma)}
  \log(1 + v\T H_{t - 1}^{-1} v)\,.
\end{align*}
\end{lemma}
\begin{proof}
We start with an upper bound on $v\T H_{t - 1}^{-1} v$. By Weyl's inequalities, we have
\begin{align*}
  \lambda_1(H_{t - 1}^{-1})
  = \lambda_d^{-1}(H_{t - 1})
  \leq \lambda_d^{-1}(\gamma I_d)
  = 1 / \gamma\,.
\end{align*}
Thus, under the assumption that $\normw{v}{2}^2 \leq c_{\max}$, we have $v\T H_{t - 1}^{-1} v \leq c_{\max} / \gamma$. Now note that for $y \in [0, y_{\max}]$,
\begin{align*}
  y
  = \frac{y}{\log(1 + y)} \log(1 + y)
  \leq \left(\max_{y \in [0, y_{\max}]} \frac{y}{\log(1 + y)}\right) \log(1 + y)
  = \frac{y_{\max}}{\log(1 + y_{\max})} \log(1 + y)\,.
\end{align*}
Finally, we set $y = v\T H_{t - 1}^{-1} v$ and $y_{\max} = c_{\max} / \gamma$, and get our claim.
\end{proof}

Now we apply \cref{ass:diverse dataset} and \cref{lem:logarithmic transformation}, use the telescoping property of the sum, and get
\begin{align*}
  \sum_{t = 1}^n v_{t, i}\T H_{t - 1}^{-1} v_{t, i}
  & \leq \kappa \sum_{t = 1}^n v_{t, I_t}\T H_{t - 1}^{-1} v_{t, I_t}
  \leq c \sum_{t = 1}^n \log(1 + v_{t, I_t}\T H_{t - 1}^{-1} v_{t, I_t})
  = c \sum_{t = 1}^n \log\det(I_d + H_{t - 1}^{- \frac{1}{2}} v_{t, I_t}
  v_{t, I_t}\T H_{t - 1}^{- \frac{1}{2}}) \\
  & = c \sum_{t = 1}^n \log\det(H_{t - 1} + v_{t, I_t} v_{t, I_t}\T) -
  \log\det(H_{t - 1})
  = c \sum_{t = 1}^n \log\det(H_t) - \log\det(H_{t - 1}) \\
  & = c (\log\det(H_n) - \log\det(H_0))
  = c \log\det(H_0^{- \frac{1}{2}} H_n H_0^{- \frac{1}{2}})\,,
\end{align*}
where $c = \frac{c_{\max} \kappa}{\gamma \log(1 + c_{\max} / \gamma)}$. Furthermore,
\begin{align*}
  \log\det(H_0^{- \frac{1}{2}} H_n H_0^{- \frac{1}{2}})
  & \leq d \log\left(\frac{1}{d}
  \trace(H_0^{- \frac{1}{2}} H_n H_0^{- \frac{1}{2}})\right)
  = d \log\left(1 + \frac{1}{d} \sum_{t = 1}^n
  \trace(H_0^{- \frac{1}{2}} v_{t, I_t}
  v_{t, I_t}\T H_0^{- \frac{1}{2}})\right) \\
  & = d \log\left(1 + \frac{1}{d} \sum_{t = 1}^n
  v_{t, I_t}\T H_0^{-1} v_{t, I_t}\right)
  \leq d \log\left(1 + \frac{c_{\max} n}{\gamma d}\right)\,.
\end{align*}
Finally, we combine all claims and get
\begin{align*}
  \phi_i\T H_n^{-1} \phi_i
  \leq \frac{1}{n} \sum_{t = 1}^n \phi_i\T H_{t - 1}^{-1} \phi_i
  \leq \frac{c_{\max} \kappa}{n}
  \sum_{t = 1}^n v_{t, I_t}\T H_{t - 1}^{-1} v_{t, I_t}
  \leq \frac{c_{\max}^2 \log\left(1 + \frac{c_{\max} n}{\gamma d}\right)}
  {\gamma \log(1 + c_{\max} / \gamma)} \frac{\kappa d}{n}\,.
\end{align*}
This completes the proof.

\section{Ablation Study}
\label{sec:ablation study}

\begin{figure*}[t!]
  \centering
  \includegraphics[width=6.4in]{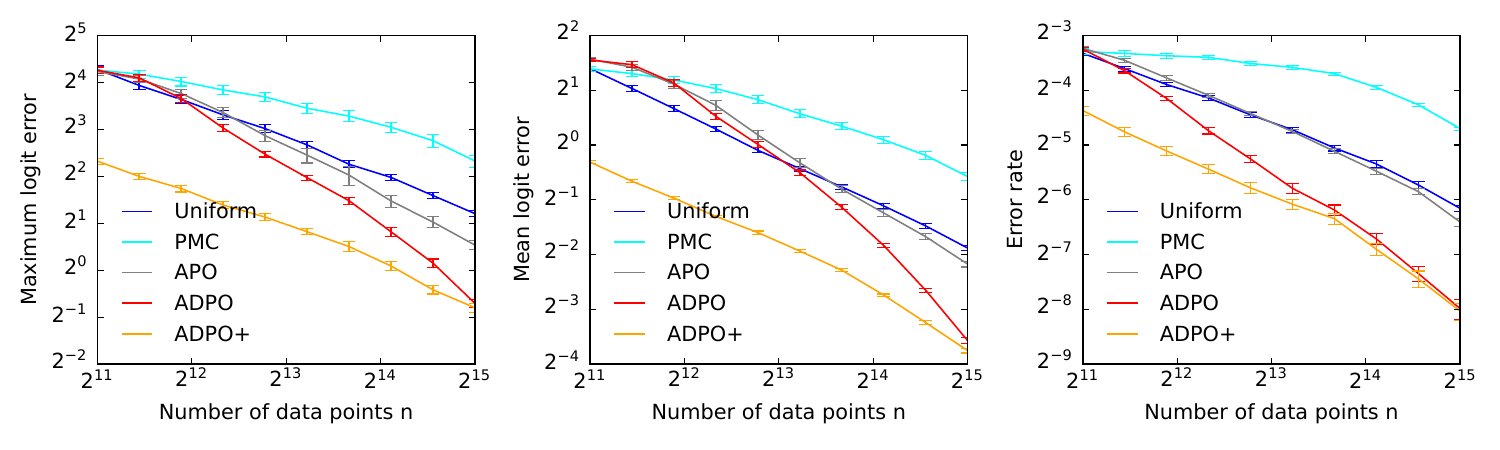}
  \vspace{-0.1in} \\
  \includegraphics[width=6.4in]{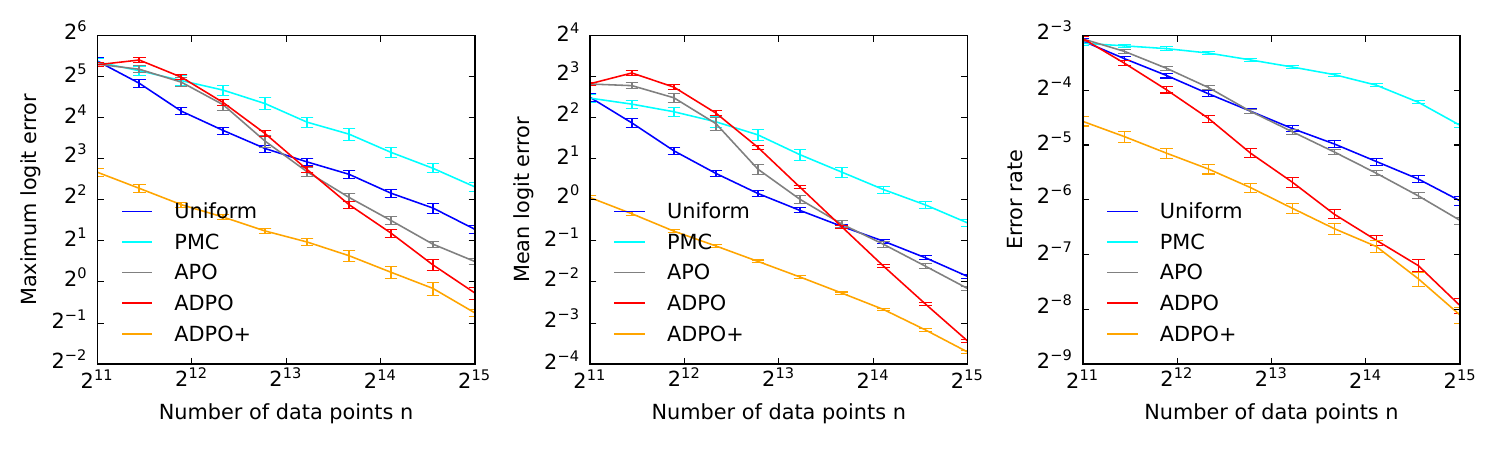}
  \vspace{-0.15in}
  \caption{Experiments with log-linear policies on the CIFAR-10 dataset, with $\beta = 2$ (first row) and $\beta = 5$ (second row).}
  \label{fig:cifar betas}
\end{figure*}

\begin{figure*}[t!]
  \centering
  \includegraphics[width=6.4in]{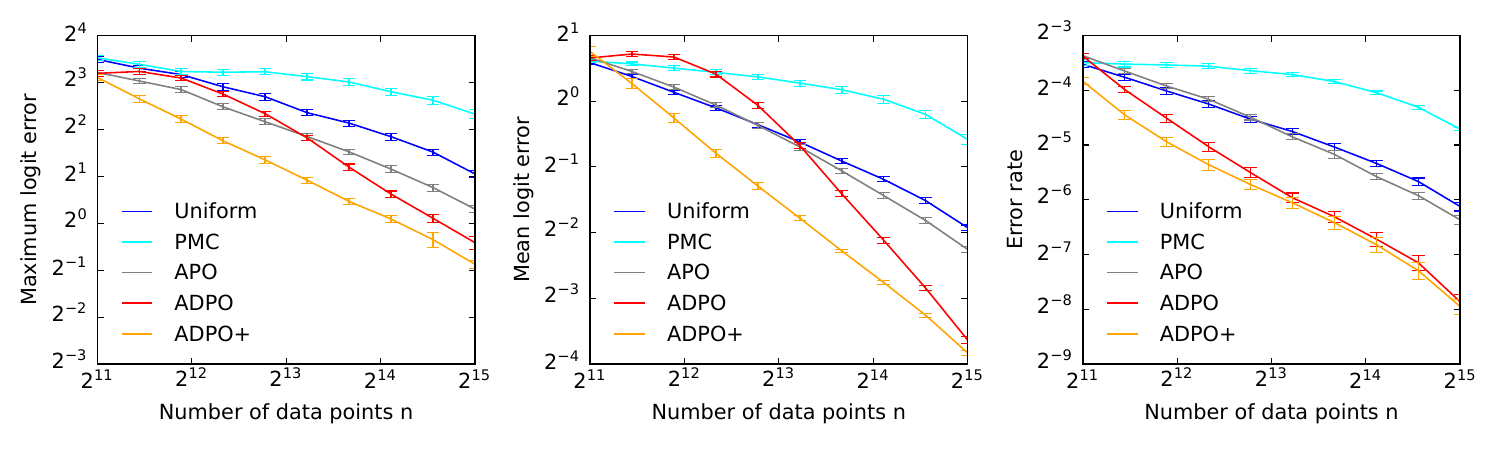}
  \vspace{-0.1in} \\
  \includegraphics[width=6.4in]{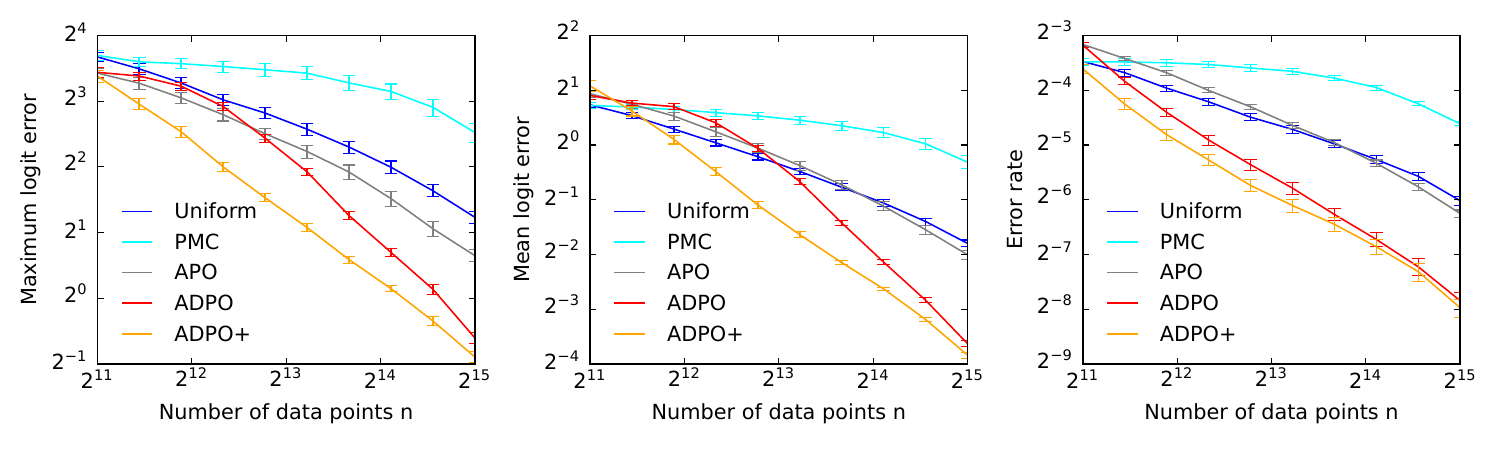}
  \vspace{-0.15in}
  \caption{Experiments with log-linear policies on the CIFAR-10 (first row) and CIFAR-100 (second row) datasets with $\alpha = 0$ in \eqref{eq:ucb}.}
  \label{fig:cifar no ucb}
\end{figure*}

In \cref{sec:log-linear policies}, we experiment with $\beta = 1$. There is nothing specific about this choice. In \cref{fig:cifar betas}, we report results for $\beta \in \set{2, 5}$ and observe improvements in both settings.

To increase the stability of our algorithms at small sample sizes, we replace $\mu_i(\hat{\theta}_t) (1 - \mu_i(\hat{\theta}_t))$ with a high probability upper confidence bound (UCB). Let $\hat{\Sigma}_t$ be the covariance matrix for $\hat{\theta}_t$. Then the UCB is computed as
\begin{align}
  U_i
  = \mu(z_i) (1 - \mu(z_i))\,, \quad
  z_i
  = \max \left\{\abs{\beta (\phi_i\T \hat{\theta}_t - b_i)} -
  \alpha \sqrt{\phi_i\T \hat{\Sigma}_t \phi_i}, \ 0\right\}
  \label{eq:ucb}
\end{align}
for some $\alpha > 0$. We set $\alpha = 3$ in \cref{sec:experiments}. In \cref{fig:cifar no ucb}, we set $\alpha = 0$ and observe that this has no major impact on our trends as the number of data points $n$ increases.

\section{Related Work}
\label{sec:related work}

The closest related works are on active learning with preferential feedback, and we review them first (\cref{sec:related work preferential feedback}). Then we review active learning for fine-tuning (\cref{sec:related work fine-tuning}) and other related works (\cref{sec:multi-armed bandits}).

\subsection{Active Learning for Preferential Feedback}
\label{sec:related work preferential feedback}

\citet{mehta23sample} applied active learning to DPO in Section 5. Their acquisition function is
\begin{align*}
  I_t
  = \argmax_{i \in [N]} (\max_{j \in [2]} U(x_i, y_{i, j}) -
  \max_{j \in [2]} L(x_i, y_{i, j}))\,,
\end{align*}
where $U(x, y)$ is the UCB and $L(x, y)$ is the LCB of $r(x, y)$. The analysis is for dueling the UCB response with a random response. Their optimized metric is the \emph{maximum gap}
\begin{align}
  \max_{i \in [N]} (\max_{j \in [2]} r(x_i, y_{i, j}) - r(x_i, \hat{y}_i))\,,
  \label{eq:maximum gap}
\end{align}
where $\hat{r}$ is the estimated reward model and $\hat{y}_i = \argmax_{j \in [2]} \hat{r}(x_i, y_{i, j})$ is the best response given $x_i$. They prove that the maximum gap is $O(1 / \sqrt{n})$ for sampling with replacement.

\citet{das24active} proposed two algorithms for active RLHF. The acquisition function in APO is
\begin{align*}
  I_t
  = \argmax_{i \in [N]} \norm{\phi_i}_{H_t^{-1}(\hat{\theta}_{t - 1})}\,,
\end{align*}
where $H_t(\hat{\theta}_{t - 1})$ is a logistic regression Hessian in round $t$, which is re-estimated in each round. They prove that \eqref{eq:maximum gap} is $O(1 / \sqrt{n})$ for sampling with replacement. APO is not evaluated. This is the closest algorithm design to \adpo. The main difference in \adpo is that we maximize the information gain (line 6) and do not compute $H_t^{-1}(\hat{\theta}_{t - 1})$. \citet{das24active} also proposed a practical APO,
\begin{align*}
  I_t
  = \argmax_{i \in [N]} \norm{\phi_i}_{H_t^{-1}}\,,
\end{align*}
where $H_t$ is a linear regression Hessian in round $t$. Practical APO is not analyzed. We use it as a baseline in \cref{sec:experiments}.

\citet{mukherjee24optimal} studied active learning with absolute and ranking feedback with $K \geq 2$ responses. For $K = 2$, their algorithm Dope is $I_t \sim \pi_*$, where $\pi_*$ is a distribution over $N$ prompts with $2$ responses obtained by the D-optimal design. They prove that
\begin{align*}
  \argmax_{i \in [N]} |\phi_i\T (\hat{\theta} - \theta_*)|
  = O(1 / \sqrt{n})
\end{align*}
for sampling with replacement, where $\theta_*$ is the true model parameter and $\hat{\theta}$ is its estimate from $n$ observations. Dope is evaluated on RLHF datasets. \citet{thekumparampil24comparing} extended \citet{mukherjee24optimal} to ranking $N$ items from $K \leq N$ responses.

\citet{liu24dual} extended APO of \citet{das24active} to selecting both the prompt and teacher model. They prove that \eqref{eq:maximum gap} is $O(1 / \sqrt{n})$ for sampling with replacement. The proposed algorithm is empirically evaluated.

\citet{scheid24optimal} proposed offline and online algorithms for active learning of reward models in RLHF. The offline algorithm, which is in the same setting as our work, computes the D-optimal design, similarly to \citet{mukherjee24optimal} for $K = 2$, and explores by sampling with replacement. They prove a $O(1 / \sqrt{n})$ bound on \eqref{eq:maximum gap}. The paper does not contain any experiments.

\citet{ji2024reinforcement} proposed two active learning algorithms: APPO and ADPO. APPO is a regret minimizing algorithm similar to those in dueling bandits. In round $t$, APPO is given a prompt as an input and proposes two responses to duel. APPO is analyzed. ADPO is a heuristic that queries responses on prompts where the agent is uncertain. The response is uncertain if $|r(x_i, y_{i, 1}) - r(x_i, y_{i, 2})|$ in the DPO objective is high.

\citet{muldrew2024active} proposed an active learning algorithm for DPO that repeatedly acquires labels and fine-tunes on them. The data are acquired in batches until a budget is met. The acquisition function is
\begin{align*}
  I_t
  = \argmax_{i \in [N]} |\hat{r}(x_i, y_{i, 1}) - \hat{r}(x_i, y_{i, 2})|\,,
\end{align*}
where $\hat{r}$ is the estimated reward model. We use it as a baseline in \cref{sec:experiments}.

\citet{guo2024direct} proposed online DPO from AI feedback. The key is to elicit AI feedback instead of human feedback and then use it in DPO. This is an empirical paper.

\citet{chen2024cost} proposed active learning with coresets for reward models. They learn cluster centroids in the space of prompt embeddings that minimize the maximum distance of the prompt to its closest centroid. This is an empirical paper.

\subsection{Active Learning for Fine-Tuning}
\label{sec:related work fine-tuning}

There are many related works on active learning in LLMs \citep{margatina2023active,bayer2024activellm,zhang2022active}. A recent survey by \citet{wang2024survey} categorizes existing methods for data selection in instruction tuning. Most of these methods rely on heuristic approaches, such as uncertainty sampling, clustering, or diversity-based strategies, which often lack theoretical grounding. \citet{doucet2024bridging} proposed a method that bridges diversity and uncertainty in active learning by leveraging self-supervised pre-training to address the cold-start problem and enhance data efficiency. However, these approaches do not align data selection directly with the task-specific objective, limiting their effectiveness in optimizing downstream performance. \citet{zhang2022active} used LLMs for selecting instances for in-context learning. More recently, \citet{bayer2024activellm} proposed ActiveLLM, which is a pool-based sampling method that leverages LLMs to select batches of instances for humans to label. Despite this fundamental difference, they also study two variants of their approach, one that incorporates feedback and another one that does not.

\subsection{Multi-Armed Bandits}
\label{sec:multi-armed bandits}

Our setting is also related to multi-armed bandits. Due to the budget $n$, it is reminiscent of fixed-budget \emph{best arm identification (BAI)} \citep{bubeck09pure,audibert10best,azizi22fixedbudget,yang22minimax}. The main difference is that we do not want to identify the best arm. We want to get a good estimate for a set of arms, essentially pairs of items, in the worst case. Online learning to rank has also been studied extensively \citep{radlinski08learning,kveton15cascading,zong16cascading,li16contextual,lagree16multipleplay}. We do not minimize cumulative regret or try to identify the best arm.

\end{document}